\documentclass[preprint,12pt]{elsarticle}

\usepackage{amssymb}
\usepackage{epsfig}
\usepackage{epsf}
\usepackage{booktabs}
\usepackage{amssymb}
\usepackage{multirow}
\usepackage{mathrsfs}
\usepackage{longtable}
\usepackage{lscape}
\usepackage{tabularx}
\usepackage{subfigure}
\usepackage{graphicx}
\usepackage{color}
\newtheorem{definition}{Definition}
\newtheorem{example}{Example}
\usepackage[perpage,symbol*]{footmisc}
\makeatletter
\def\@makefnmark{}
\makeatother

\journal{Journal of Systems Engineering and Electronics}
\begin{document}

\begin{frontmatter}



\title{A new combination approach based on improved evidence distance}
\author[label1,label5]{Hongming Mo}
\author[label1,label3,label4]{Yong Deng\corref{cor1}}
\ead{ydeng@swu.edu.cn, prof.deng@hotmail.com}
\cortext[cor1]{Corresponding author at: School of Computer and Information Science, Southwest University, Chongqing 400715, China. Tel.: +86 023 68254555; Fax: +86 023 68254555.}

\address[label1]{School of Computer and Information Science, Southwest University, Chongqing 400715,China}
\address[label5]{Department of the Tibetan Language, Sichuan University of Nationalities, Kangding  Sichuan 626001, China}
\address[label3]{School of Electronics and Information Technology, Shanghai Jiao Tong University, Shanghai  200240, China}
\address[label4]{School of Engineering, Vanderbilt University, TN 37235, USA }
\begin{abstract}
 Dempster-Shafer evidence theory is a powerful tool in information fusion. When the evidence are highly conflicting, the  counter-intuitive results will be presented. To adress this open issue, a new method based on evidence distance of Jousselme and Hausdorff distance  is proposed. Weight of each  evidence can be computed, preprocess the original evidence to generate a new evidence. The Dempster's  combination rule is used to combine the new evidence. Comparing with the existing methods, the new proposed method is efficient.
 \end{abstract}

\begin{keyword}
Evidence theory\sep
Conflict evicence\sep
Evidence distance \sep
Combination  rule \sep
Target recognition

\end{keyword}
\end{frontmatter}
 \footnote{This work is partially supported by National Natural Science Foundation of China (Grant No. 61174022), National High Technology Research and Development Program of China(863 Program) (No. 2013AA013801), Chongqing Natural Science Foundation (Grant No. CSCT, 2010BA2003), the Chenxing Scholarship Youth Found of Shanghai Jiao Tong University (Grant No.T241460612). }


\section{Introduction}
Dempster-Shafer evidence theory\cite{dempster1967upper,shafer1976mathematical} has attracted more and more attentions recently years. It  can handle with uncertain and incomplete information in many fields, such as target recognition, information fusion and decision making\cite{denoeux2008conjunctive,dubois1988representation,heyounewmethod2011,han2011weighted1,han2011weighted,zhang2000new,pan2001some,he2012new,deng2011risk,deng2011new,deng2011new1,deng2010target,suo2013computational,tan2012data,geng2013consensus,wei2013identifying,gao2013modified,kang2012evidential,chen2013fuzzy}.
While the evidence are highly conflicting, the  Dempster's combination rule  will generate counter-intuitive results, such as the  typical conflictive  example proposed by Zadeh\cite{zadeh1986simple}. In the last decade researchers have proposed many  approaches to cope with this open issue and certain effort have been obtained. The existing methods can be mainly classified into two categories.  The first  strategy  regards that Dempster's combination rule is incomplete and modifying the  combination rule as alternative, such as Yager's method\cite{yager1987dempster}, Smet's method\cite{smets1994transferable,smets1990combination} and  Lefevre's method\cite{lefevre2002belief}, etc. The second  strategy believes that  Dempster's rule has perfect theoretical foundation and preprocessing the original evidence before combination, such as Haenni's method\cite{haenni2002alternatives}, Murphy's method\cite{murphy2000combining} and  Deng's method\cite{deng2004efficient}, etc. We believe that Dempster's rule is excellent and has been widely  applied in recent years. In this paper, preprocessing the original evidence for highly conflicting  is adopted. The method of Deng proposed\cite{deng2004efficient} in 2004   based on the evidence distance can  deal with the conflicting evidence and that the correct sensor can be quickly recognized. The evidence distance of Deng's  method reflects the difference between evidences distance roughly, but can not reflect  the degree of difference. In this paper, we propose a new method weighted averaging the evidence, improving Deng's  method\cite{deng2004efficient}. The new method takes both  Jousselme\cite{jousselme2001new}  and Hausdorff\cite{hausdorff1957set} evidence distance into account. Thus, the weights of evidence are more appropriate.

The remainder of  this paper is organized as follows. Section 2 presents some preliminaries. The proposed method is presented in section 3. Numerical examples and applications are used to demonstrate the validity of the proposed method in section 4. A short conclusion is drawn in the last section.
\section{Preliminaries}
In this section, some concepts of Dempster-Shafer evidence theory\cite{dempster1967upper,shafer1976mathematical}  are briefly recalled. For more information please consult Ref.\cite{he2010information}. The Dempster-Shafer evidence theory  is introduced by Dempster and then developed by Shafer.

In Dempster-Shafer evidence theory, let $\Theta =\left\{ {\theta _1 ,\theta _2 , \cdots ,\theta _n} \right\}$
be the finite set of mutually exclusive and exhaustive elements.
It is concerned with the set of all subsets of $\Theta$, which is a powerset of $2^{\left| \Theta  \right|}$, known as the frame of discernment, denotes as
\[\Omega  = \left\{ {\emptyset ,\left\{ {{\theta _1}} \right\},\left\{ {{\theta _2}} \right\},\left\{ {{\theta _3}} \right\}, \cdots ,\left\{ {{\theta _n}} \right\},\left\{ {{\theta _1},{\theta _2}} \right\}, \cdots ,\left\{ {{\theta _1},{\theta _1} \cdots ,{\theta _n}} \right\}} \right\}\]

 The mass function of evidence assigns probability to the subset of $\Omega$, also called basic probability assignment(BPA), which satisfies the following conditions
\[m(\phi ) = 0,0 \le m(A) \le 1,\sum\limits_{A \subseteq \theta } {m(A) = 1}. \]
 $\phi$ is an empty set and $A$ is any subsets of $\Theta$.

Dempster's combination rule\cite{dempster1967upper,shafer1976mathematical}  is the first one within the framework of evidence theory which can combine two BPAs $m_1$ and $m_2$ to yield a new BPA $m$. The  rule of Dempster's combination is presented as follows
\begin{equation}
m(A) = \frac{1}{{1 - k}}\sum\limits_{B \cap C = A} {{m_1}(B){m_2}(C)}
\end{equation}
with
\begin{equation}
k = \sum\limits_{B \cap C = \emptyset } {m_1 (B)m_2 (C)}
\end{equation}

Where  $k$ is a normalization constant, namely the conflict coefficient of  BPAs.

\section{New combination approach}

The method of Murphy\cite{murphy2000combining} purposed  regards each BPA as the same role, little relevant to the relationship among the BPAs. In Deng's weighted  method\cite{deng2004efficient}, each BPAs play different roles, that depended on the extent to which they are accredited in system. The similarity  of Deng's method between two BPAs is ascertained by Jousselme distance function\cite{jousselme2001new}.
\subsection{Two existing evidence distance}
The evidence distance proposed by Jousselme\cite{jousselme2001new} is presented as follows
\begin{definition}
Let $m_1$ and $m_2$ be two BPAs defined on the same frame of discernment  $\Theta$, containing $N $ mutually exclusive and exhaustive hypotheses. The metric $d_{BPA}$ can be defined as follows

\begin{equation} \label{dpa123}
d_{BPA} (m_1 ,m_2 ) = \sqrt {\frac{1}{2}\left( {{m_1 }  - {m_2 } } \right)^T \underline{\underline D} \left( { {m_1 }  -  {m_2 } } \right)}
\end{equation}
\end{definition}
${\underline{\underline D} } $ is a  $2^N \times 2^N $ similarity matrix, indicates  the conflict of focal element in $m_1$ and $m_2$, where
\begin{equation}\label{DJ}
 \underline{\underline D} (A,B) = \frac{{|A \cap B|}}{{|A \cup B|}}
 \end{equation}

$\left| {A \cup B} \right|$ is the cardinality of subset of the union $A$ and $B$, where $A$ and $B$ may belong to the same BPA or  come from different BPAs. $\left| {A \cap B} \right|$ indicates the conflict degree  between  elements $A$ and  $B$. When two elements have no common object,  they are highly conflicting.

Another evidence distance proposed by Sunberg\cite{sunberg2013belief} is presented as follows
\begin{definition}
Let $m_1$ and $m_2$ be two BPAs defined on the same frame of discernment $\Theta$, containing $N $ mutually exclusive and exhaustive hypotheses. The distance of  two BPAs  referred to as $d_{Haus}$ is defined as follows
\begin{equation}
d_{Haus} (m_1 ,m_2 ) = \sqrt {\frac{1}{2}\left( {{m_1 }  - \ {m_2 } } \right)^T {D_{H}} \left( {{m_1 }  - {m_2 } } \right)}
\end{equation}
\end{definition}
with
\begin{equation}\label{Dijhaus}
D_{H(i,j)}=S_{H}(A_{i},A_{j})=\frac{1}{1+CH(A_{i},A_{j})}
\end{equation}

Where H($A_{i}$,$A_{j}$) is the Hausdorff distance\cite{hausdorff1957set} between focal elements $A_{i}$ and $A_{j}$. $A_{i}$ and $A_{j}$ may  belong to the same BPA or come form different  BPAs.  Positive number $C$ is a user-defined tuning parameter. $C$ is set to be 1, in this paper, for simplicity. 
It is defined according to
 \begin{equation} \label{DHaus}
{H}(A_{i},A_{j})=\max\{\sup_{b \subseteq A_{i}}\inf_{c \subseteq A_{j}}d(b,c), \sup_{c \subseteq A_{j}}\inf_{b \subseteq A_{i}}d(b,c)\}
\end{equation}

 Where $d(b,c)$ is the distance between two elements of the sets and can be defined as any valid metric distance on the measurement space\cite{hausdorff1957set}.

While the elements are real numbers, the Hausdorff distance may be simplify as\cite{hausdorff1957set,sunberg2013belief}
 \begin{equation}\label{DHaus1}
H_{R}(A_{i},A_{j})=\max\{|\min(A_{i})-\min(A_{j})|, |\max(A_{i})-\max(A_{j})|\}
\end{equation}

The below example is used to illustrate the difference between Jousselme  distance\cite{jousselme2001new} and Hausdorff distance\cite{hausdorff1957set}.
\begin{example}
There are five orderable mutually exclusive and exhaustive hypotheses elements: 1, 2, 3, 4 and 5 on the same frame of discernment $\Theta$.
\end{example}

By (\ref{DJ}), the Jousselme  distance matrix  $\underline{\underline D}$ between each elements in a BPA can be  obtained as follows
\[\underline{\underline D} = \left[ {\begin{array}{*{20}{c}}
1&0&0&0&0\\
0&1&0&0&0\\
0&0&1&0&0\\
0&0&0&1&0\\
0&0&0&0&1
\end{array}} \right]\]

Utilize Hausdorff distance in (\ref{Dijhaus}), the Hausdorff distance matrix $D_H$ between each elements in a BPA can be obtained  as follows
\[{D_H} = \left[ {\begin{array}{*{20}{c}}
1&{\frac{1}{2}}&{\frac{1}{3}}&{\frac{1}{4}}&{\frac{1}{5}}\\
{\frac{1}{2}}&1&{\frac{1}{2}}&{\frac{1}{3}}&{\frac{1}{4}}\\
{\frac{1}{3}}&{\frac{1}{2}}&1&{\frac{1}{2}}&{\frac{1}{3}}\\
{\frac{1}{4}}&{\frac{1}{3}}&{\frac{1}{2}}&1&{\frac{1}{2}}\\
{\frac{1}{5}}&{\frac{1}{4}}&{\frac{1}{3}}&{\frac{1}{2}}&1
\end{array}} \right]\]

It is clearly that, the five elements have no object in common.  The similarity between each elements  are the same value zero in Jousselme  distance matrix. In case of this, Jousselme  distance matrix can not show the detailed distance of each elements in an orderable system. However, Hausdorff distance matrix can calculate  the detail  similarity between each orderable elements.

\subsection{New combination approach}
In this subsection, we purpose an improved  combination approach based on Deng's method\cite{deng2004efficient}. The new method takes advantage of  Huasdorff  distance\cite{hausdorff1957set} to update  Jousselme distance\cite{jousselme2001new}.
\begin{definition}
Let $m_1$ and $m_2$ be two BPAs defined on the same frame of discernment $\Theta$, containing $N $ mutually exclusive and exhaustive hypotheses. The distance between $m_1$ and $m_2$ can be defined as
\begin{equation}\label{newmethod}
d_{Com} (m_1 ,m_2) = \sqrt {\frac{1}{2}\left( {{m_1 }  - \ {m_2 } } \right)^T {D_{Com}} \left( {{m_1 }  - {m_2 } } \right)}
\end{equation}
\end{definition}
with
\begin{equation}\label{d}
{D_{Com}{(i,j)}} = {{\underline{\underline D}}{(i,j)}}.*{D_H}{(i,j)}\end{equation}

$D_{Com}$ is a  $2^N \times 2^N $ similarity matrix, indicates  the metric of focal elements in $m_1$ and $m_2$.  $\underline{\underline D}{(i,j)}$ is the distance matrix in  (\ref{DJ}) and ${D_H}{(i,j)}$ is the distance matrix in  (\ref{Dijhaus}).

Given there are $n$ BPAs in the system, we can calculate the distance between each two BPAs. Thus, the distance matrix is presented as follows
\begin{equation}\label{dm}
DIM = \left[ {\begin{array}{*{20}{c}}
1&{{d_{12}}}& \cdots &{{d_{1j}}}& \cdots &{{d_{1n}}}\\
 \vdots & \vdots & \vdots & \vdots & \vdots & \vdots \\
{{d_{i1}}}&{{d_{i2}}}& \cdots &{{d_{ij}}}& \cdots &{{d_{in}}}\\
 \vdots & \vdots & \vdots & \vdots & \vdots & \vdots \\
{{d_{n1}}}&{{d_{n2}}}& \cdots &{{d_{nj}}}& \cdots &1
\end{array}} \right]
\end{equation}

\begin{definition}
Let $Simi({m_i},{m_j})$ be the similarity value between BPA $m_{i}$ and $m_{j}$, thus the $Simi({m_i},{m_j})$ can be defined as

\begin{equation}
Simi({m_i},{m_j}) = 1 - {d_{Com}}({m_i},{m_j})
\end{equation}
\end{definition}

It is obvious that while the value of distance between two BPAs are bigger, the similarity of two BPAs are smaller, and vice versa. The similarity function can be represented by a matrix as follows
\begin{equation}\label{sim}
SIM = \left[ {\begin{array}{*{20}{c}}
1&{Simi_{12}}& \cdots &{Simi_{1j}}& \cdots &{Simi_{1n}}\\
 \vdots & \vdots & \vdots & \vdots & \vdots & \vdots \\
{Sim{i_{i1}}}&{Sim{i_{i2}}}& \cdots &{Sim{i_{ij}}}& \cdots &{Sim{i_{in}}}\\
 \vdots & \vdots & \vdots & \vdots & \vdots & \vdots \\
{Sim{i_{n1}}}&{Sim{i_{n2}}}& \cdots &{Sim{i_{nj}}}& \cdots &1
\end{array}} \right]
\end{equation}

\begin{definition}
Let $Supp(m_i)$ be the support degree of BPA $m_i$ in the system, and the support degree of BPA $m_i$ can be presented as follow
\begin{equation}\label{supp}
Supp({m_i}) = \sum\limits_{\scriptstyle j = 1\hfill \atop \scriptstyle j \ne i\hfill}^n {Simi({m_i},{m_j})}
\end{equation}
\end{definition}

 From  (\ref{sim}) and   (\ref{supp}), we can see that the support degree $Supp({m_i})$ is the sum of similarity  between each  BPAs, except itself. The  larger the value of $Supp({m_i})$ is, the  more important the evidence will be.

To normalize  $Supp(m_i)$, the  $W(m_i)$ of BPA $m_i$ can be obtained as follows
\begin{equation}\label{weight}
W({m_i}) = \frac{{Supp({m_i})}}{{\sum\limits_{i = 1}^n {Supp({m_i})} }}\end{equation}
It is obvious that
\[\sum\limits_{i = 1}^n {W({m_i}) = 1} \]

$W({m_i})$ indicates the important and credible degree of  BPA $m_i$ among all BPAs in the system. It can be regard as the weight of BPA $m_i$. After obtained the weight  of each BPAs, we take advantage of Dempster's combination rule\cite{dempster1967upper,shafer1976mathematical} to yield a new BPA.

 The below example is  used to demonstrate the detail processes of the new proposed method.
\begin{example}
Given there  are four BPAs $m_1$, $m_2$, $m_3$ and $m_4$ on the same frame of discernment $\Theta$:
\[\begin{array}{l}
{m_1}(R) = 0.3,{m_1}(S) = 0.5,{m_1}(T) = 0.2\\
{m_2}(R) = 0,{m_2}(S) = 0.5,{m_2}(T) = 0.5\\
{m_3}(R) = 0.6,{m_3}(S) = 0.2,{m_3}(T) = 0.2\\
{m_4}(R) = 0.9,{m_4}(S) = 0,{m_4}(T) = 0.1
\end{array}\]
\end{example}

By (\ref{newmethod})-(\ref{weight}), we can obtain the weight of the four BPAs  $m_1$, $m_2$, $m_3$ and $m_4$ as follows
\[W({m_1}) = 0.2688,W({m_2}) = 0.2276,W({m_3}) = 0.2752,W({m_4}) = 0.2284.\]
Therefore, the new BPA  $m_{New}$ before combination can be obtained as follows
\[\begin{array}{l}
{m_{New}}(R) = 0.3\times0.2688 + 0\times0.2276 + 0.6\times0.2752 + 0.9\times0.2284 = 0.4513\\
{m_{New}}(S) = 0.5\times0.2688 + 0.5\times0.2276 + 0.2\times0.2752 + 0\times0.2284 = 0.3033\\
{m_{New}}(T) = 0.2\times0.2688 + 0.5\times0.2276 + 0.2\times0.2752 + 0.1\times0.2284 = 0.2454
\end{array}\]

There are four BPAs in this example, we apply Dempster's combination rule to combine the new BPA  $m_{New}$ three times, the results are presented as follows
 \[m(R) = 0.7744,m(S) = 0.1579,m(T) = 0.0677.\]

\section{Numerical examples and Applications}

It is known that Dempster-Shafer evidence theory\cite{dempster1967upper,shafer1976mathematical} needs less information than Bayes  probability to deal with uncertain information. It  is often  regarded as the extension  of  Bayes  probability.

We utilize the below example  to illustrate the effectiveness of the new proposed method.
\begin{example}
There are   five mass functions on the same frame of discernment, the five BPAs are presented as follows\cite{deng2004efficient}
\[\begin{array}{l}
{m_1}:{m_1}(A) = 0.5,{m_1}(B) = 0.2,{m_1}(C) = 0.3\\
{m_2}:{m_2}(A) = 0,{m_2}(B) = 0.9,{m_2}(C) = 0.1\\
{m_3}:{m_3}(A) = 0.55,{m_3}(B) = 0.1,{m_3}(C) = 0.35\\
{m_4}:{m_4}(A) = 0.55,{m_4}(B) = 0.1,{m_4}(C) = 0.35\\
{m_5}:{m_5}(A) = 0.55,{m_5}(B) = 0.1,{m_5}(C) = 0.35
\end{array}\]
\end{example}

The results of different methods to combine the five BPAs are presented in Table.\ref{tabel1}. From Table.\ref{tabel1}, we can see that Dempster's combination rule\cite{dempster1967upper,shafer1976mathematical} can not handel with highly conflicting evidence. Once an element is negatived by  any  BPAs,  no matter how strongly it is supported by other BPAs, its probability  will  always remain zero.

\begin{figure}[!h]
\begin{center}
\psfig{file=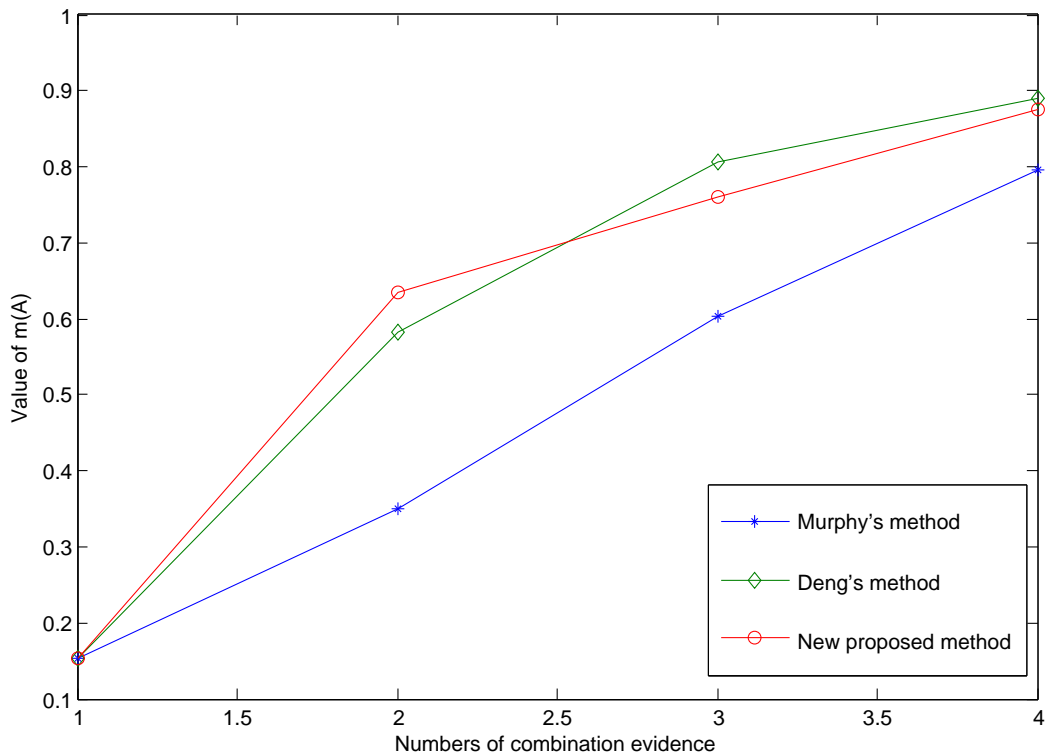,scale=0.8}
\caption{The convergence speeds of  different approaches.}
\label{speed}
\end{center}
\end{figure}

Murphy's method\cite{murphy2000combining} regards each  evidence plays the same role in the system, considered little relations among evidences. Deng\cite{deng2004efficient} improved Murphy's work and took advantage of an evidence distance as the weight of each evidence. The novel proposed method  based on Deng's method, but utilizes Hausdorff distance  to update the distance matrix. Fig.\ref{speed} indicates that the convergence speed of  proposed method is  slower than Deng's method but faster than Murphy's method, owing to the additional update distance because some sensors may be orderable.

\begin{table}[htbp]
\centering
\caption{Different combination rules to combine highly conflicting evidence.}
\begin{center}
\begin{tabular}{cllll} \toprule
\addtolength\doublerulesep{1pt}
\addtolength{\tabcolsep}{1ex}
 &$m_1,m_2$&$m_1,m_2,m_3$&$m_1,m_2,m_3,m_4$&$m_1,m_2,m_3,m_4,m_5$\\\hline
Dempster's&$m(A)=0$&$m(A)=0$&$m(A)=0$&$m(A)=0$\\
combination&$m(B)=0.8571$&$m(B)=0.6316$&$m(B)=0.3288$&$m(B)=0.1228$\\
rule\cite{dempster1967upper,shafer1976mathematical}&$m(C)=0.1429$&$m(C)=0.3684$&$m(C)=0.6712$&$m(C)=0.8772$\\
\quad&\quad&\quad&\quad&\quad\\

Murphy's &$m(A)=0.1543$&$m(A)=0.3500$&$m(A)=0.6027$&$m(A)=0.7958$\\
combination&$m(B)=0.7469$&$m(B)=0.5224$&$m(B)=0.2627$&$m(B)=0.0932$\\
 rule\cite{murphy2000combining}&$m(C)=0.0988$&$m(C)=0.1276$&$m(C)=0.1346$&$m(C)=0.1110$\\
\quad&\quad&\quad&\quad&\quad\\
Deng's      &$m(A)=0.1543$&$m(A)=0.5816$&$m(A)=0.8060$&$m(A)=0.8909$\\
combination&$m(B)=0.7469$&$m(B)=0.2439$&$m(B)=0.0482$&$m(B)=0.0086$\\
            rule\cite{deng2004efficient}&$m(C)=0.0988$&$m(C)=0.1745$&$m(C)=0.1458$&$m(C)=0.1005$\\
\quad&\quad&\quad&\quad&\quad\\
New proposed     &$m(A)=0.1543$&$m(A)=0.6355$&$m(A)=0.7605$&$m(A)=0.8761$\\
combination&$m(B)=0.7469$&$m(B)=0.2229$&$m(B)=0.0897$&$m(B)=0.0189$\\
            rule&$m(C)=0.0988$&$m(C)=0.1415$&$m(C)=0.1468$&$m(C)=0.1050$\\
\toprule
\end{tabular}
\end{center}
\label{tabel1}
\end{table}

\section{Conclusion}
Dempster-Shafer evidence theory is a powerful tool to deal with uncertain and imprecise information in widely fields. However the evidence collected  may be multifarious, some of them may be highly conflicting, owing to various noise factors, subjective or objective.  The original Dempster combination rule can do nothing for these highly conflicting evidence. Modified  methods of Dempster's combination rule are briefly introduced, and all of them have some drawbacks. The new proposed method inherits all the advantages of Deng's method. It  applies Hausdorff distance to update the Jousselme  distance and takes more distance information  into account. Numerical examples demonstrate  that the new proposed method can discern the correct target, effectively.

\bibliographystyle{elsarticle-num}
\bibliography{coresample}
\section*{Biographies}
\textbf{Hongming Mo} was born in 1983. He received the B.S.degree from Chongqing Normal University in 2006. He is now an assistant researcher in Sichuan University of Nationalities. His research interests include uncertain information modeling and processing.

\textbf{Yong Deng }was born in 1975. He received the Ph.D.degree from Shanghai Jiaotong University in 2003. He is now a professor in Southwest University. His research interests include uncertain information modeling and processing.

\end{document}